\documentclass[table]{bmvc2k}
\usepackage{paralist} 
\usepackage{enumitem} 
\usepackage{graphicx}
\usepackage{amsmath}
\usepackage{resizegather}
\usepackage{amssymb}
\usepackage{amsfonts}
\usepackage{booktabs}
\usepackage{siunitx}
\usepackage{mathtools}
\usepackage{fixltx2e} 
\newcommand{\NA}{---}
\usepackage{graphics} 
\usepackage{amssymb} 
\usepackage{wrapfig} 
\usepackage{array} 
\usepackage{xcolor}
\definecolor{shadecolor}{RGB}{255,203,203}

\usepackage{pifont}
\title{Revisiting spatio-temporal layouts for compositional action recognition}

\addauthor{Gorjan Radevski}{gorjan.radevski@esat.kuleuven.be}{12}
\addauthor{Marie-Francine Moens}{sien.moens@cs.kuleuven.be}{2}
\addauthor{Tinne Tuytelaars}{tinne.tuytelaars@esat.kuleuven.be}{1}

\addinstitution{
 PSI-ESAT,\\
 KU Leuven,\\
 Leuven, Belgium
}
\addinstitution{
 LIIR-CS Department,\\
 KU Leuven,\\
 Leuven, Belgium
}

\runninghead{Radevski et al.}{Revisiting spatio-temporal layouts}

\begin{document}

\setlength{\abovedisplayskip}{-9pt}
\setlength{\belowdisplayskip}{2pt}

\maketitle

\begin{abstract}
Recognizing human actions is fundamentally a spatio-temporal reasoning problem, and should be, at least to some extent, invariant to the appearance of the human and the objects involved. Motivated by this hypothesis, in this work, we take an object-centric approach to action recognition. Multiple works have studied this setting before, yet it remains unclear
\begin{inparaenum}[(i)]
\item how well a carefully crafted, spatio-temporal layout-based method can recognize human actions, and
\item how, and when, to fuse the information from layout- and appearance-based models.
\end{inparaenum}
The main focus of this paper is compositional/few-shot action recognition, where we advocate the usage of multi-head attention (proven to be effective for spatial reasoning) over spatio-temporal layouts, i.e., configurations of object bounding boxes. We evaluate different schemes to inject video appearance information to the system, and benchmark our approach on background cluttered action recognition. On the Something-Else and Action Genome datasets, we demonstrate
\begin{inparaenum}[(i)]
\item how to extend multi-head attention for spatio-temporal layout-based action recognition,
\item how to improve the performance of appearance-based models by fusion with layout-based models,
\item that even on non-compositional background-cluttered video datasets, a fusion between layout- and appearance-based models improves the performance.
\end{inparaenum}
\end{abstract}

\section{Introduction}

Whether a person is "taking an apple out of a box" or  "taking a screwdriver out of a box", we can recognize the action performed with ease. In fact, even if we have never seen the object before, we are still able to recognize the action that occurred. Moreover, for us, it makes no difference where the action takes place (indoors, outdoors, etc.), as long as the objects involved in the action are visible. This suggests that action recognition should be, to a degree, invariant to the appearance of the objects, as well as the environment where the action takes place. Yet, most state-of-the-art action recognition methods are appearance-based 3D CNNs \cite{lin2019tsm, yang2020temporal, xie2018rethinking, carreira2017quo}. These methods are indeed powerful, albeit heavily reliant on large-scale (pre)training datasets \cite{kataoka2020would}. Unfortunately, in spite of the great pre-training efforts taken, their performance rapidly deteriorates on compositional action recognition, i.e., when the objects encountered at test time are novel \cite{materzynska2020something}.\par
For this reason, multiple other works \cite{baradel2018object, wang2018videos, sun2018actor, girdhar2019video, zhang2019structured} have advocated an object-centric approach for video action recognition, reporting an improved robustness, interpretability, and overall performance. To achieve object-centric reasoning, on top of the video appearance (RGB frames), these works either use a region proposal network \cite{wang2018videos, girdhar2019video, baradel2018object, sun2018actor}, or leverage an independently trained object detector, e.g., Faster R-CNN \cite{ren2015faster}, to obtain object detections for each video frame \cite{materzynska2020something, yan2020interactive}. Within these models, the \textit{layout} module operates on object detections, while the \textit{appearance} module operates on RGB frames. In a \textit{two-branch model} the layout and appearance input follow separate pathways and are fused late, while in a \textit{one-branch model} they follow a single pathway (fused early in the model). Some of the limitations include:
\begin{inparaenum}[(i)]
\item With a two-branch model, fusion is performed by concatenation, not fully exploiting the complementarity of the spatio-temporal layouts and the video appearance \cite{baradel2018object, materzynska2020something, wang2018videos}, and
\item the layout module is treated as a peripheral component \cite{wang2018videos, ji2020action}, so it remains unclear to what extent in different evaluation settings (compositional, few-shot, background cluttered videos), a well assembled layout-based model can recognize human actions.
\end{inparaenum}
At the same time, a multi-head attention model \cite{vaswani2017attention} has been demonstrated to be a powerful common-sense reasoning tool over sets of spatially distributed objects in images for visual question-answering \cite{tan2019lxmert, lu2019vilbert}, layout generation \cite{radevski2020decoding}, etc. By applying multiple heads of beyond-pairwise spatial reasoning, it encapsulates the scene's global spatial context, which is indicative of its semantics, to a certain extent. Just as importantly, a variety of works specifically examine the problem of multimodal fusion \cite{neverova2015moddrop,vielzeuf2018centralnet,perez2019mfas}, attempting to determine how and where to fuse the different modalities.\par
\textbf{Contributions.} The main focus of this paper is compositional and few-shot action recognition, where we hold on to the object-level video reasoning and 
\begin{inparaenum}[(i)]
\item reveal how a multi-head attention based method, applied purely over highly abstract concepts (no appearance information), i.e., spatio-temporal layouts, can be extended for action recognition, 
\item investigate how to fuse the information from the layout- and appearance-based branch for improved action recognition,
\item find that, even on non-compostional, background cluttered video dataset such as Action Genome \cite{ji2020action}, reasoning over the spatio-temporal layouts significantly improves the performance. The codebase and trained models are released \href{https://github.com/gorjanradevski/revisiting-spatial-temporal-layouts}{here}\footnote{\href{https://github.com/gorjanradevski/revisiting-spatial-temporal-layouts}{https://github.com/gorjanradevski/revisiting-spatial-temporal-layouts}}.
\end{inparaenum}

\section{Related work}
Action recognition methods are mostly 3D CNN based \cite{carreira2017quo, xie2018rethinking, lin2019tsm, simonyan2014two, kataoka2020would, wang2016temporal, simonyan2014two, feichtenhofer2019slowfast, yang2020temporal}. These methods often use a 2D CNN pre-trained on ImageNet \cite{deng2009imagenet}, subsequently inflated to 3D \cite{carreira2017quo}. Other works explore (pre)training 3D CNNs \cite{kataoka2020would} on large-scale curated datasets \cite{carreira2019short, monfort2019moments}, as well as the best practices for doing so \cite{wang2016temporal}, reducing the computational complexity \cite{lin2019tsm, xie2018rethinking, tran2018closer}, or propose plug-in components to improve the temporal reasoning \cite{zhou2018temporal}.\par

\textbf{Multi-head attention (MHA) in computer vision.} The applications of MHA in computer vision are rapidly expanding \cite{khan2021transformers}. So far, MHA has been applied in conjunction with a CNN \cite{carion2020end, sun2019videobert, girdhar2019video, meng2019interpretable}, as a stand-alone MHA over low level, raw image pixels \cite{dosovitskiy2021an, wang2020axial, Cordonnier2020On, bertasius2021space}, for vision + text tasks \cite{lu2019vilbert, tan2019lxmert}, or tasks involving spatial reasoning \cite{radevski2020decoding} to name a few. In contrast, we apply MHA:
\begin{inparaenum}[(i)]
\item over high level, abstract, spatio-temporal layouts, and
\item to fuse the features of two distinct modalities (layout and appearance).
\end{inparaenum}
\par

\textbf{Object level reasoning for action recognition (with attention).} The issue with appearance-based action recognition methods is their inherent tendency to overfit on the appearance of the environment and of the objects, deteriorating the performance on fine-grained \cite{baradel2018object} or compositional datasets \cite{materzynska2020something}. Recently, multiple works that address this issue emerged \cite{baradel2018object, materzynska2020something, girdhar2019video, ji2020action, wang2018videos, sun2018actor, perez2020knowing, xu2020spatio, materzynska2020something, yan2020interactive}. Object Relation Network \cite{baradel2018object} performs spatio-temporal reasoning over detected video objects with a GRU \cite{cho-etal-2014-learning}. STAR \cite{xu2020spatio} regresses the bounding box coordinates where the action occurs and classifies the action. STRG \cite{wang2018videos} uses a graph CNN \cite{kipf2017semi} and a region proposal network (RPN) \cite{ren2015faster}, applied over I3D \cite{carreira2017quo} features, with a non-local neural network (NL) \cite{wang2018non} temporal module. Actor Centric Relation Network \cite{sun2018actor} fuses the cropped feature map of the actors' regions and the global video feature map. In parallel, multiple works leverage attention, to augment existing methods or propose new ones. LFB \cite{wu2019long} uses attention as a non-local block \cite{wang2018non} to accumulate video features. SINet's \cite{ma2018attend} coarse- and fine-grained branch are attention-based, subsequently fused for action recognition. Compared to us, SINet's fine-grained branch applies attention over the region of interest (RoI) pooled features from an RPN for each frame, subsequently fed to LSTM \cite{hochreiter1997long}, while we apply MHA over the object detections (category + bounding box) to encode the videos' spatio-temporal context. W3 \cite{perez2020knowing} is an attention based plug-in module on top of appearance models,  while SGFB \cite{ji2020action} utilizes attention through LFB \cite{wu2019long} over per-frame scene graphs, combined with I3D \cite{carreira2017quo} and NL \cite{wang2018non}. The Video Action Transformer (VAT) \cite{girdhar2019video} uses I3D \cite{carreira2017quo} in conjunction with RPN \cite{ren2015faster} and a transformer \cite{vaswani2017attention}. It
\begin{inparaenum}[(i)]
\item obtains an I3D feature map around a center frame,
\item generates region proposals for the center frame,
\item applies MHA where the I3D feature map is the memory and the center frame RoI pooled features are the query.
\end{inparaenum}
In our work, we also benchmark a VAT inspired fusion scheme between the appearance and layout branch. Lastly, STIN \cite{materzynska2020something} and SFI \cite{yan2020interactive} demonstrate that a graph neural network \cite{kipf2017semi} layout model can surpass I3D's \cite{carreira2017quo} performance for compositional/few-shot action recognition with ground truth object detections, and fusion with I3D improves performance. Unlike these works, inspired by MHA-based methods for spatial reasoning, we
\begin{inparaenum}[(i)]
\item develop a specifically tailored model for layout-based action recognition which applies attention over high-level, spatio-temporal layouts,
\item empirically evaluate different state-of-the-art MHA-based fusion methods, to uncover how and when the appearance- and layout-based models should be fused.\par
\end{inparaenum}
\textbf{Multimodal fusion} attempts to extract the relevant, complementary information from multimodal input, resulting in a better joint model, compared to training separate models on the individual modalities. The literature is extensive \cite{lahat2015multimodal, vielzeuf2018centralnet, neverova2015moddrop, arevalo2017gated}, without a universal approach that generalizes across different modalities and tasks. Many works \cite{tan2019lxmert, lu2019vilbert} that use MHA have demonstrated remarkable results on multimodal tasks, e.g., VQA \cite{antol2015vqa}, when fusing image and text features. In action recognition, late fusion by concatenation works well \cite{materzynska2020something}, also confirmed in our work. We demonstrate that multimodal fusion with cross-attention \cite{tan2019lxmert}, based on the CentralNet approach \cite{vielzeuf2018centralnet}, further improves the performance.\par

\section{Methodology}
Next, we introduce the necessary multi-head attention background (Sec.~\ref{sec:revisit}), describe how it is extended for modelling spatio-temporal layouts, i.e., object detections (Sec.~\ref{sec:spatial-temporal}), and discuss the appearance models and  schemes  to fuse them with the layout model~(Sec.~\ref{sec:appearance-fusion}).

\subsection{Multi-head attention revisited}\label{sec:revisit}
The core Transformer model component \cite{vaswani2017attention} is the multi-head attention module, defined as: $\operatorname{Attention}(Q, K, V) = \operatorname{Softmax}(\frac{Q\cdot K^T}{\sqrt{d_k}})\cdot V$, where $Q$, $K$, $V$ are the queries, keys and values respectively, and $d_k$ is the per-attention head hidden size. With self-attention, $Q$, $K$ and $V$ come from the same and only modality (in our case, the layout modality), and with cross-attention, $Q$ originates from the target, while $K$ and $V$ from the source modality we attend on. Due to MHA's permutation invariance, the input should include positional information. In this work, we rely on
\begin{inparaenum}[(i)]
\item bidirectional and causal attention,
\item self- and cross-attention, and
\item different variants of position embeddings.
\end{inparaenum}
Refer to \cite{vaswani2017attention} for details.
\begin{figure*}[t]
\centering
\includegraphics[width=0.8\textwidth]{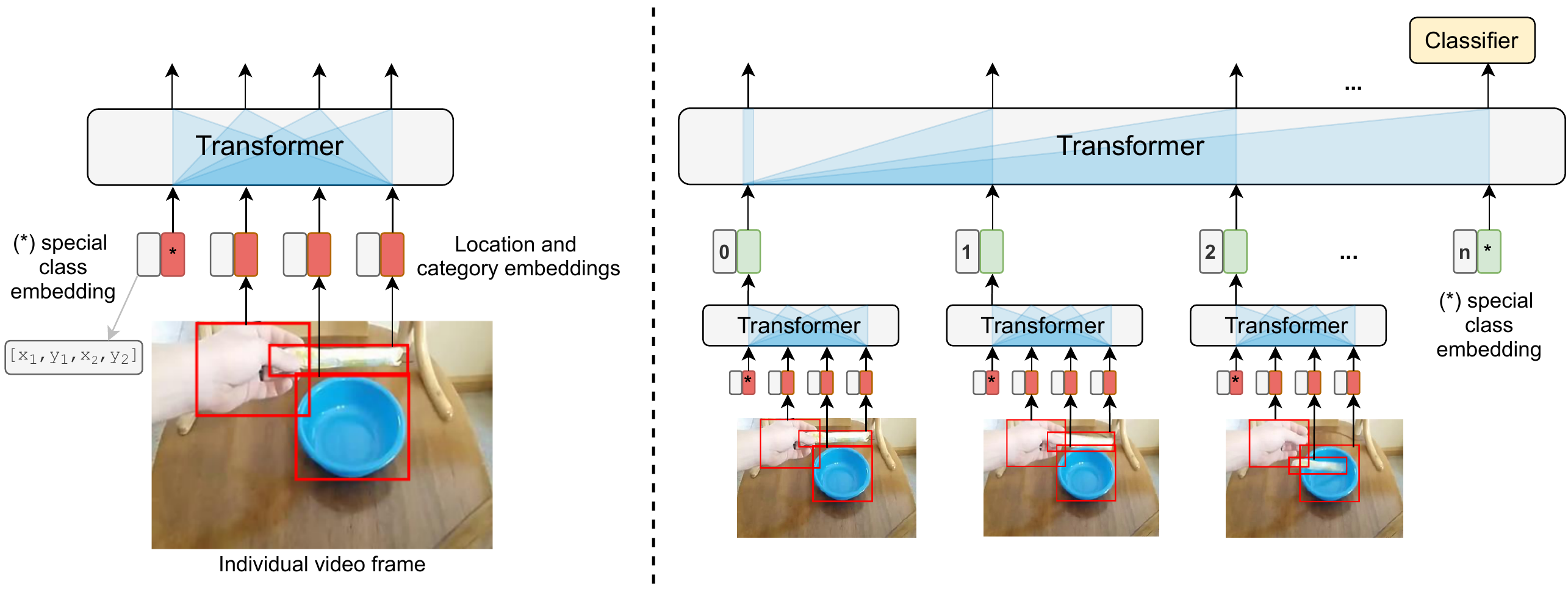}
\caption{STLT overview. \textbf{Left: Spatial Transformer}. The inputs are the object categories, and their [$x_1, y_1, x_2, y_2$] frame location normalized by the frame size. We select the special \texttt{class} embedding as the module output. \textbf{Right: Temporal Transformer}. The inputs are the Spatial Transformer outputs, summed with trainable position embeddings. We select the temporal \texttt{class} embedding as the module output, and add a classifier for action recognition.}
\label{fig:main-model}
\end{figure*}

\subsection{Layout branch: Spatial-Temporal Transformer}\label{sec:spatial-temporal}
The input to the layout model (Fig.~\ref{fig:main-model}) is a frame sequence $S = (f_0, f_1,...,f_{n-1})$ of length~$n$. Each frame $f_i$ is composed of $m$ objects, $f_i = \{o_0, o_1,...,o_{m-1}\}$, where $o_j$ consists of the object category $c_j$ and location in the frame $l_j = [x_1, y_1, x_2, y_2]$. \textit{Note that this is all the information required and used by the layout branch: no appearance information, just bounding boxes and category labels.}
Two separate fully-connected layers yield the category embedding $\hat{c}_j$ and the frame location embedding $\hat{l}_j$, which we subsequently sum together and apply layer-normalization~\cite{ba2016layer} and dropout \cite{srivastava2014dropout} to obtain the final object embedding: $\hat{o}_j = \operatorname{Dropout(LayerNorm(\hat{c}_j + \hat{l}_j))}$.
\par
With the layout model, dubbed as Spatial-Temporal Layout Transformer,
henceforth abbreviated as STLT, 
we decouple the spatial (per-frame) reasoning from the temporal (across-video) reasoning. To that end, to model the per-frame spatial relations, given a set of frame objects $f_i = \{o_0, o_1, ..., o_{m-1}\}$, we firstly prepend an $o_{\text{\texttt{class}}}$ object, with $c_{\text{\texttt{class}}}$ (special \texttt{class} category)
and $l_{\text{\texttt{class}}}$ equal to the frame size. Then, we obtain the embedding $\hat{o}_j$ for each frame object, $\hat{f}_i = \{\hat{o}_{\text{\texttt{class}}}, \hat{o}_0, \hat{o}_1, ..., \hat{o}_{m-1}\}$, and use a bidirectional transformer (each object embedding can attend on all others). We denote this module as $\operatorname{Spatial-Transformer}$ (Fig.~\ref{fig:main-model}, left), which we apply on the set of object embeddings for each frame $\hat{f}_i$ separately. Subsequently we select the output hidden state corresponding to the \texttt{class} category as a global representation of a single frame: $\hat{s}_i = \operatorname{Spatial-Transformer}(\hat{f}_i)$.\par
To model the temporal evolution of the spatial relations, we use a causal transformer (each frame embedding can attend on the past ones), denoted as $\operatorname{Temporal-Transformer}$ (Fig.~\ref{fig:main-model}, right). We firstly append another special \texttt{class} embedding $\hat{s}_{\text{\texttt{class}}}$ to the outputs of the Spatial-Transformer. Then, for each frame, with a fully-connected layer, we obtain frame-position-in-the-video embedding $\hat{p}_i$. This is summed with the frame spatial embedding $\hat{s}_i$, followed by layer-normalization and dropout: $\hat{t}_i = \operatorname{Dropout(\operatorname{LayerNorm(\hat{s}_i + \hat{p}_i)})}$. 
Finally, we forward propagate the sequence of frame embeddings $\hat{T} = (\hat{t}_0, \hat{t}_1, ... \hat{t}_{n-1}, \hat{t}_{\text{\texttt{class}}})$ through the Temporal Transformer: $\hat{H} = \operatorname{Temporal-Transformer}(\hat{T})$, where $\hat{H}$ are the output hidden states. If we use STLT as a standalone action recognizer, i.e., given spatio-temporal layouts as input we want to infer the action, we select the hidden state corresponding to the \texttt{class} embedding, and add a classifier on top: $\hat{y} = \operatorname{Linear}(\hat{h}_{\text{\texttt{class}}})$, where $\hat{y}$ are the logits.

\subsection{Appearance branch and multimodal fusion}\label{sec:appearance-fusion}
\begin{figure*}[t]
\centering
\includegraphics[width=0.9\textwidth]{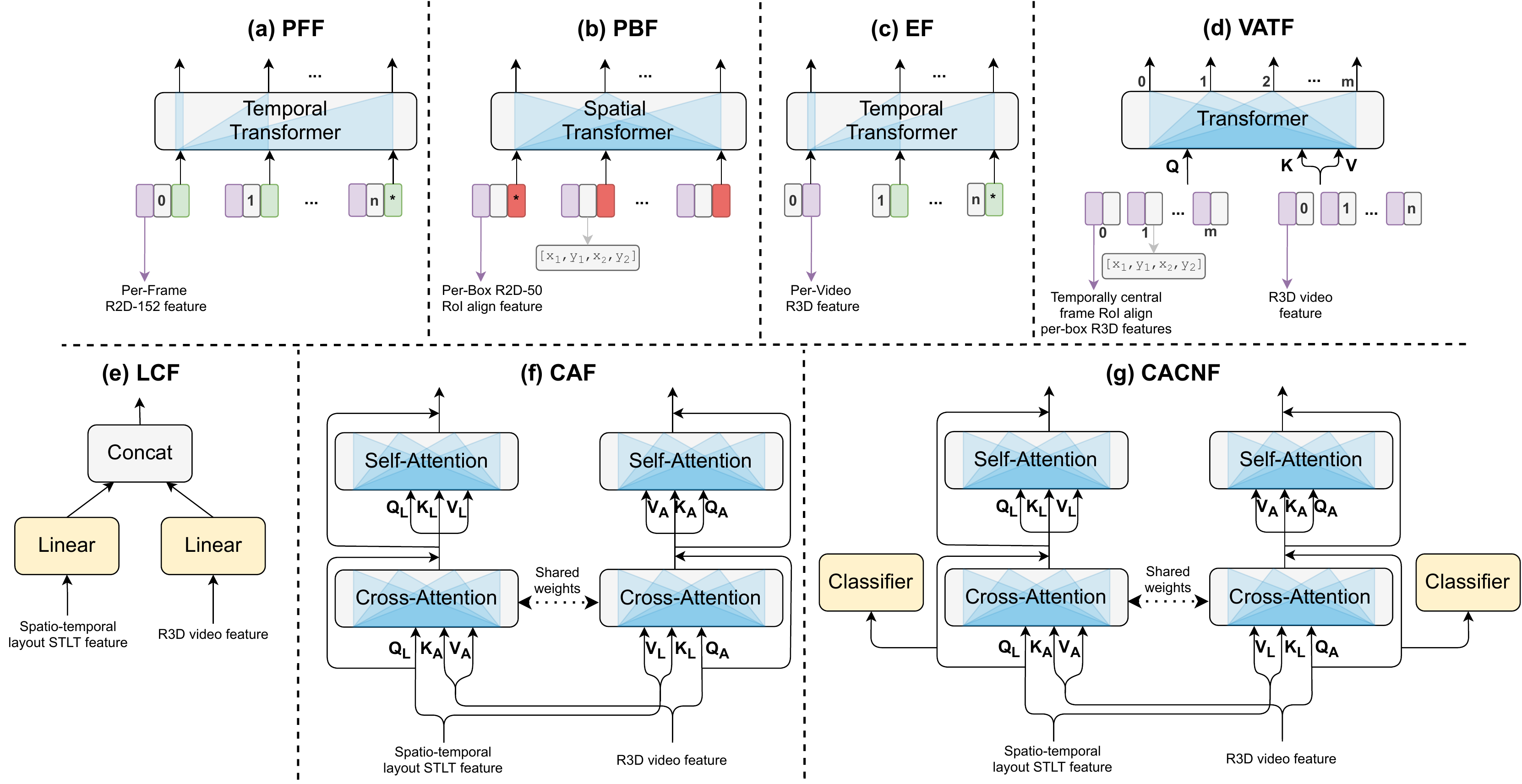}
\caption{Fusion schemes. \textbf{Top:} One-branch. \textbf{Bottom:} Two-branch fusion methods.}
\label{fig:fusion-overview}
\end{figure*}
As an appearance model, we deem a neural network, usually a CNN \cite{lecun1998gradient}, applied over pixels of the video frame(s). In this work, we use four types of appearance features, each tightly coupled with the corresponding fusion approaches:
\begin{inparaenum}[(i)]
\item 2D Resnet152 (R2D-152) \cite{he2016deep}, pre-trained on ImageNet \cite{deng2009imagenet}, applied over individual frames;
\item 2D Resnet50 (R2D-50) backbone,
from a COCO \cite{lin2014microsoft} pre-trained Faster R-CNN \cite{ren2015faster}. Given class-agnostic bounding boxes from a video frame,
we extract RoI align \cite{he2017mask} features from each;
\item An inflated 3D Resnet50 (R3D) \cite{kataoka2020would}, pre-trained on \cite{carreira2019short, monfort2019moments, yoshikawa2018stair};
\item R3D, same as (iii), with a Transformer encoder to enable multimodal fusion with cross-attention.\par
\end{inparaenum}
Due to the specific nature of the appearance features, we devise different ways to fuse them with the layout model (Fig.~\ref{fig:fusion-overview}). We evaluate different configurations where each has its weaknesses, while we gradually build the ultimate, empirically superior fusion approach (for details see supplementary):
\begin{inparaenum}[(a)]
\item \textbf{Per-Frame Fusion (PFF) --} We sum the R2D-152 and the Spatial Transformer features for each frame individually, and feed the obtained embedding as input to the Temporal Transformer;
\item \textbf{Per-Box Fusion (PBF) --} We sum the RoI aligned R2D-50 features with the bounding box and the category embeddings, and feed the fused embedding as input to the Spatial Transformer;
\item \textbf{Early Fusion (EF) --} We feed the R3D video features as a first token to the Temporal Transformer, essentially allowing each spatial frame embedding to attend on the entire video;
\item \textbf{Video Action Transformer Fusion (VATF) --} A fusion approach which is an adapted Video Action Transformer (VAT) model \cite{girdhar2019video} for our task. We use the R3D as a trunk, and leverage the externally obtained bounding boxes to extract RoI align features from the temporally central frame as a query, and the trunk features as a memory to a transformer model which predicts the action (for details refer to \cite{girdhar2019video});
\item \textbf{Late Concatenation Fusion (LCF) --} The R3D video embedding is concatenated with the STLT embedding right before the classifier, commonly used as a standard baseline for multimodal fusion;
\item \textbf{Cross-Attention Fusion (CAF) --} A multimodal fusion with cross-attention \cite{tan2019lxmert} to fuse the layout (STLT) and appearance (R3D) branch embeddings;
\item \textbf{Cross-Attention CentralNet Fusion (CACNF) --} A CentralNet \cite{vielzeuf2018centralnet} based CAF implementation. Despite performing end-to-end training by minimizing the loss from the cross-attention fusion module (CAF) output, we additionally minimize the loss for each branch (layout and appearance) independently. To that end, by using the CentralNet fusion approach, we achieve multimodal fusion on a two-branch model, where the individual branches are enforced to preserve their individual abilities.
\end{inparaenum}

\section{Evaluation and discussion}\label{sec:eval}

\begin{wraptable}{R}{7cm}
\vspace{-15pt}
\begin{center}
\resizebox{0.5\textwidth}{!}{
\begin{tabular}{ccccccc} \toprule
    {} & \multicolumn{4}{c}{Something-Else: Compositional setting} & \multicolumn{2}{c}{Something-Something} \\ 
    {} & \multicolumn{2}{c}{Obj. predictions} & \multicolumn{2}{c}{Oracle} & \multicolumn{2}{c}{Oracle}  \\ \midrule
    {Method} & {Top 1 acc.} & {Top 5 acc.}  & {Top 1. acc.} & {Top 5 acc.} & {Top 1 acc.} & {Top 5 acc.} \\ \midrule
    GNN-NL & 33.3 & 58.9 & 50.7 & 78.6 & 47.1 & 76.3 \\
    S\&TLT & 40.6 & 66.9 & 57.7 & 84.7 & 55.6 & 84.3 \\
    STLT & \textbf{41.6} & \textbf{67.7} & \textbf{59.4} & \textbf{85.8} & \textbf{57.0} & \textbf{85.2} \\
     \midrule
     R3D & 51.3 & 78.6 & 51.3 & 78.6 & 52.2 & 80.7 \\
     \midrule
    PFF & 47.3 & 73.7 & 62.5 & 87.5 & 62.9 & 88.0 \\
    PBF & 48.5 & 73.2 & 62.9 & 86.5 & \textbf{64.5} & 89.1 \\
    EF & \textbf{52.8 }& \textbf{79.3} & \textbf{63.8} & \textbf{88.1} & 64.4 & \textbf{89.4} \\
    VATF & 49.1 & 78.0 & 53.0 & 79.6 & 54.9 & 82.8 \\ \midrule
    LCF & 54.1 & 79.8 & 66.1 & 88.8 & 64.4 & 89.3 \\
    CAF & 52.3 & 78.9 & 64.4 & 88.6 & 64.5 & 89.1 \\
    \rowcolor{shadecolor}
    CACNF & \textbf{56.9} & \textbf{82.5} & \textbf{67.1} & \textbf{90.4} & \textbf{66.8} & \textbf{90.6} \\
    \bottomrule
\end{tabular}
}
\end{center}
\caption{Comparison between the different model configurations. \textbf{From top to bottom:} Layout-based methods, R3D, one-branch fusion methods, two branch fusion methods. Best method within group in bold, overall best method in red.}
\label{table:ablation-studies}
\end{wraptable}

We perform experiments on the Something-Something \cite{goyal2017something} / Else \cite{materzynska2020something} and the Action Genome datasets \cite{ji2020action}. During training, we randomly sample 16 frames (each represented as spatio-temporal layouts) for STLT, and uniformly sample 32 RGB frames for models using R3D, subsequently rescaled to 112 $\times$ 112 (complete experimental setup in supplementary).\par
\textbf{Something-Something V2} \cite{goyal2017something} consists of egocentric videos of people performing actions with their hands, with 174 unique actions. To deal with the environment bias, videos recorded by the same person can be in either the training or validation set. Nevertheless, the objects the person interacts with might still overlap between training and test time, indicating that appearance-based models can overfit on the objects' appearance.\par
\textbf{Something-Else} \cite{materzynska2020something} proposes two data splits according to the objects' distribution at training and test time.
In the compositional split, to validate the compositional generalization, the data is divided such that the models encounter distinct objects during training and testing. The training and validation set contain $\sim$55k and $\sim$58k videos respectively, with 174 actions. In the few-shot split, there are $\sim$112k pre-training videos (with 88 base actions), 5 $\times$ 86 and 10 $\times$ 86 videos in the 5-shot and 10-shot setup respectively for fine-tuning, and $\sim$49k testing videos (with 86 novel actions)\footnote{When fine-tuning, we freeze the backbone's weights and only train the action classifier following \cite{materzynska2020something}.}. On the compositional and few-shot splits, we perform experiments with Faster R-CNN \cite{ren2015faster} object detections (object predictions setting), and with ground truth object detections (oracle setting), both released by \cite{materzynska2020something}. The input object categories in STLT are either ``hand'' or ``object''. We measure performance using top-1 and top-5 accuracy (acc.), and perform training with cross-entropy loss.\par
\textbf{Action Genome} \cite{ji2020action}, built on top of Charades \cite{sigurdsson2016hollywood}, has $\sim$10k videos of people doing daily activities. Multiple object-specific actions simultaneously occur in each video out of 157 unique ones. The frames where the action occurs, i.e., the person interacts with the objects, are annotated with bounding boxes and categories. We train a Faster R-CNN \cite{ren2015faster} on these frames and obtain object detections (for details see supplementary). We perform experiments on the Charades train/validation split with our object detections (obj. predictions setting), as well as the ground truth object detections (oracle setting) released by \cite{ji2020action}.
We measure performance using mean average precision (mAP), and perform training with binary cross-entropy loss.\par

\subsection{Ablation studies and main findings}
\begin{wrapfigure}{R}{6.8cm}
\centering
\vspace{-7pt}
\includegraphics[width=0.45\textwidth]{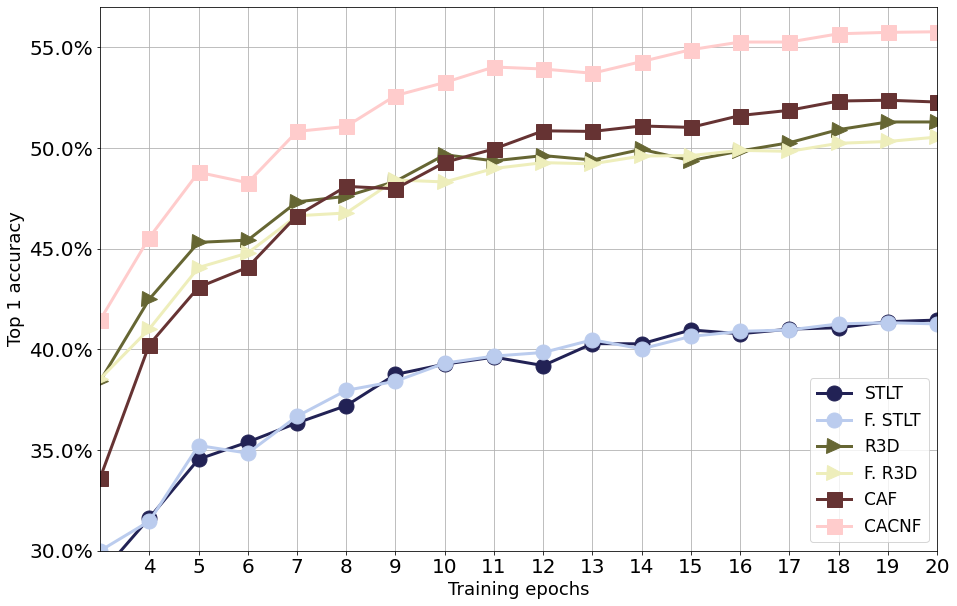}
\caption{Top-1 validation acc. (epoch 3 to 20) of STLT and R3D trained individually, trained in conjunction with CACNF,  CAF and CACNF.}
\label{fig:caf-cacnf}
\end{wrapfigure}
We ablate our models to gain insights in \textit{why} one should leverage spatio-temporal layouts for action recognition, and \textit{how} to come up with the best approach for it. We do an ablation study on the Something-Else compositional dataset, while we also verify our findings on the Something-Something (non-compositional) validation set, albeit only in an oracle setting.\par
\textbf{Layout branch: How to model the spatio-temporal layouts?} In Table~\ref{table:ablation-studies} (Top), we measure STLT's performance against:
\begin{inparaenum}[(i)]
\item A baseline model \cite{materzynska2020something} with a spatial reasoning graph neural network (GNN) and temporal reasoning non-local block (NL);
\item An STLT variant performing joint spatial-temporal reasoning (S\&TLT) on unrolled frames' bounding boxes (for details see supplementary).
\end{inparaenum}
Across different settings and layout types (obj. predictions or oracle), 
we observe that a decoupled spatio-temporal reasoning is preferable. Furthermore, STLT and S\&TLT significantly outperform GNN-NL, suggesting the appropriateness of MHA-based methods for modelling spatio-temporal layouts.\par
\textbf{Multimodal fusion: How and where to fuse?} In Table~\ref{table:ablation-studies} (middle, bottom) we report action recognition results with the fusion methods we consider in this work. We compare among the fusion methods plus a fine-tuned R3D \cite{kataoka2020would}, and succinctly summarize our empirical findings as:
\begin{inparaenum}[(i)]
\item 2D fusion methods (PFF, PBF) exhibit good performance on non-compositional datasets (Something-Something), where object appearance matters, while their performance deteriorates on compositional datasets;
\item Early fusion (EF) yields a competitive performance across different datasets and is superior to the other one-branch fusion methods;
\item The Video Action Transformer \cite{girdhar2019video} fusion type (VATF), does not fully exploit the spatial-temporal layout and video-context specific to the action (it only performs RoI align on the temporally central frame), thus it yields consistently lower results on these types of data;
\item One-branch methods only marginally outperform R3D without oracle layouts;
\item Late fusion by concatenation (LCF) remains a strong baseline, as reported by others \cite{vielzeuf2018centralnet, neverova2015moddrop};
\item Cross-Attention Fusion (CAF) is consistently weaker than LCF and CACNF, while CACNF (CAF with CentralNet \cite{vielzeuf2018centralnet}) outperforms other methods regardless of the data type (compositional, non-compositional) and type of layouts (obj. predictions or oracle).\par
\end{inparaenum}

\begin{wrapfigure}{R}{6.8cm}
\centering
\vspace{-2pt}
\includegraphics[width=0.37\textwidth]{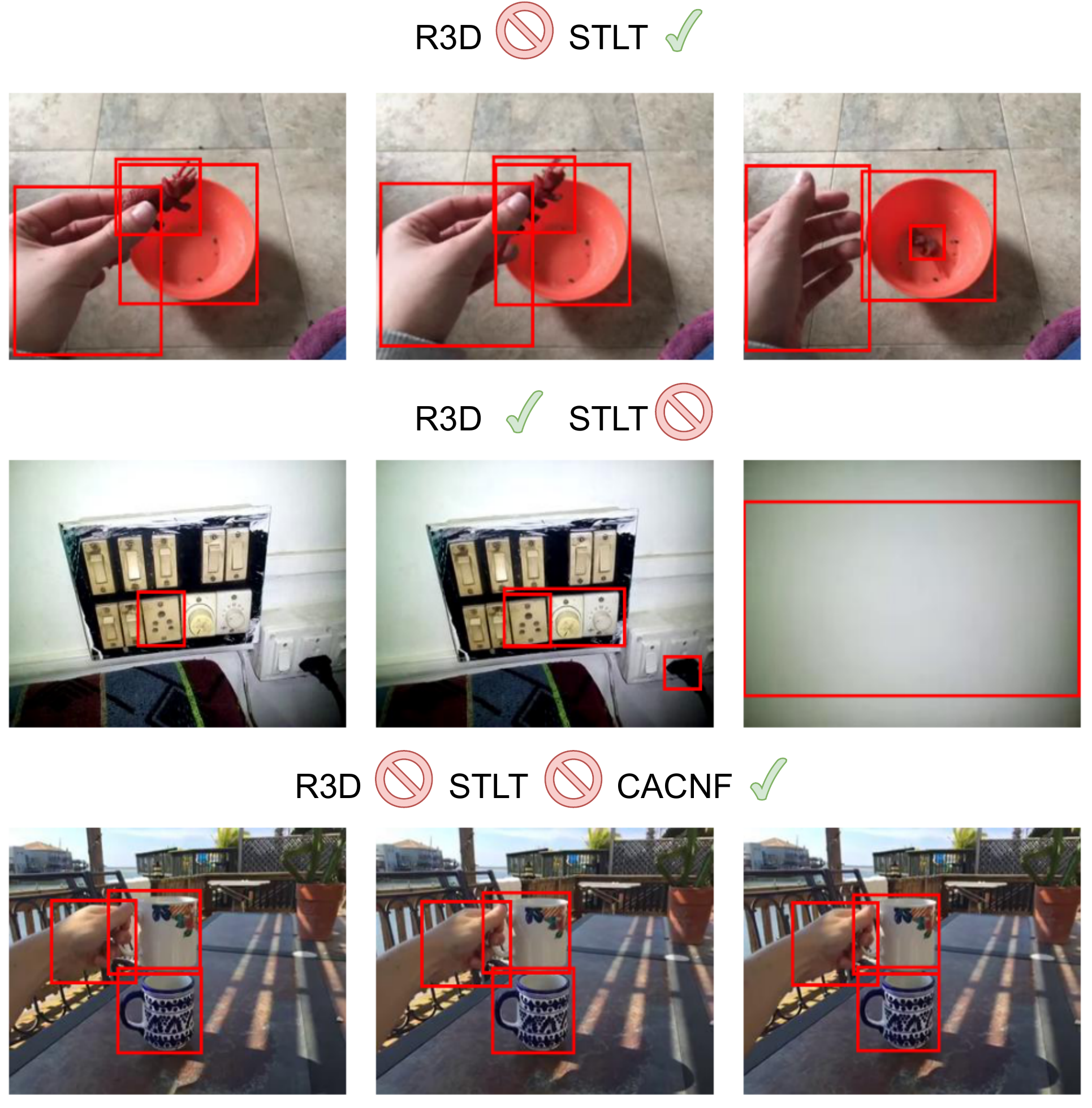}
\caption{\textbf{Top:} R3D mispredicts, STLT predicts correctly. \textbf{Middle:} STLT mispredicts, R3D predicts correctly. \textbf{Bottom:} STLT and R3D mispredict, CACNF predicts correctly.} 
\vspace{-5pt}
\label{fig:qualitative}
\end{wrapfigure}

\textbf{Why is CACNF superior to CAF despite the conceptual similarity?} In Fig.~\ref{fig:caf-cacnf}, we observe the top 1 acc. on the Something-Else compositional split, in the obj. predictions setting, of:
\begin{inparaenum}[(i)]
\item STLT trained individually;
\item STLT trained within the CACNF model (F. STLT);
\item R3D trained individually;
\item R3D trained within the CACNF model (F. R3D);
\item CAF;
\item CACNF.
\end{inparaenum}
We see that the performance of STLT and F. STLT, and, R3D and F. R3D, is remarkably similar, indicating that with CACNF the layout and appearance branch preserve their individual capabilities. What is interesting is that this phenomenon results in better overall performance of the cross-attention fusion module (CACNF), compared to training it without CentralNet \cite{vielzeuf2018centralnet} -- CAF.\par
\begin{wrapfigure}{R}{6.8cm}
\centering
\vspace{-1pt}
\includegraphics[width=0.47\textwidth]{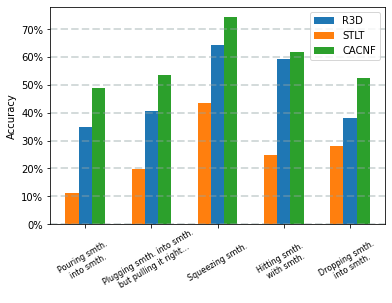}
\caption{Five Something-Else actions where the performance difference between R3D and STLT (averaged) with CACNF is most prominent.} 
\vspace{-5pt}
\label{fig:qualitative-barplot}
\end{wrapfigure}

\textbf{How does it look visually?} We are interested in visually inspecting three error types:
\begin{inparaenum}[(i)]
\item R3D predicts wrong, STLT predicts correct action;
\item STLT predicts wrong, R3D predicts correct action;
\item STLT and R3D predict wrong, CACNF predicts correct action.
\end{inparaenum}
In Fig.~\ref{fig:qualitative} (top), we observe the action ``Dropping smth. into smth.'', suitable for layout-based methods, e.g., STLT, as they directly model the spatial properties of the objects (location, movement, size, etc.). On the contrary, STLT is unable to recognize the action ``Turning the camera upwards while filming smth.'', Fig.~\ref{fig:qualitative} (middle), due its spatial ambiguity, while R3D recognizes the change in appearance, indicative of the action. Lastly, the action ``Holding smth. over smth.'' in Fig.~\ref{fig:qualitative} (bottom), requires modelling both the temporal consistency of the layout and appearance, which STLT and R3D individually fail, while CACNF recognizes the correct action. Furthermore, in Fig.~\ref{fig:qualitative-barplot}, we compare the performance of R3D and STLT with CACNF, on five actions from Something-Else, where the difference between the averaged R3D and STLT accuracy, and CACNF accuracy for each action is most prominent. We observe a consistent pattern across all five actions, i.e., CACNF successfully fuses the appearance (R3D) and layout branch (STLT), and it yields superior performance compared to the unimodal (layout or appearance) methods.

\subsection{Something-Else: State-of-the-art comparisons}

We compare against the following methods:
\begin{inparaenum}[(i)]
\item \textbf{I3D} \cite{carreira2017quo}: An inflated Resnet50 \cite{he2016deep} based 3D CNN as in \cite{wang2018videos}, pre-trained on ImageNet \cite{deng2009imagenet}, subsequently fine-tuned on the Something-Else dataset;
\item \textbf{STRG} \cite{wang2018videos}: A multimodal method, with a GNN
\cite{kipf2017semi} applied over region proposals, combined with I3D using late fusion;
\item \textbf{STIN} \cite{materzynska2020something}: A GNN \cite{kipf2017semi} for spatial, and a non-local neural network \cite{wang2018non} for temporal reasoning;
\item \textbf{STIN + I3D}: STIN combined with I3D in a late-fusion manner;
\item \textbf{SFI} \cite{yan2020interactive}: A layout-appearance fusion method,
trained with an auxiliary task of predicting the future state of the video objects.
\end{inparaenum}

\begin{table*}[t]
\begin{center}
\resizebox{\textwidth}{!}{
\begin{tabular}{ccccccccc} \toprule
    {} & \multicolumn{4}{c}{Compositional setting} & \multicolumn{4}{c}{Few-shot setting} \\
    {} & \multicolumn{2}{c}{Obj. predictions} & \multicolumn{2}{c}{Oracle} & \multicolumn{2}{c}{Obj. predictions} & \multicolumn{2}{c}{Oracle}  \\ \midrule
    {Method} & {Top 1 acc.} & {Top 5 acc.}  & {Top 1. acc.} & {Top 5 acc.} & {Top 1 acc. (5-shot)} & {Top 1. acc. (10-shot)} & {Top 1 acc. (5-shot)} & {Top 1 acc. (10-shot)}\\ \midrule
    STIN \cite{materzynska2020something} & 37.2 & 62.4 & 51.4 & 79.3 & 17.7 & 20.8 & 27.7 & 33.5 \\
    SFI \cite{yan2020interactive} & \NA & \NA & 44.1 & 74.0 & \NA & \NA & 24.3 & 29.8 \\
    STLT (Ours) & \textbf{41.6} & \textbf{67.9} & \textbf{59.0} & \textbf{86.0} & \textbf{18.8} & \textbf{24.8} & \textbf{31.4} & \textbf{38.6} \\ \midrule
    I3D \cite{carreira2017quo} & 46.8 & 72.2 & 46.8 & 72.2 & 21.8 & 26.7 & 21.8 & 26.7 \\
   STIN \cite{materzynska2020something} + I3D \cite{carreira2017quo} & 48.2 & 72.6 & 54.6 & 79.4 & 23.7 & 27.0 & 28.1 & 33.6 \\
   STRG \cite{wang2018videos} & 52.3 & 78.3 & \NA & \NA & 24.8 & 29.9 & \NA & \NA \\
    SFI \cite{yan2020interactive} & \NA & \NA & 59.6 & 85.8 & \NA & \NA & 30.7 & 36.2 \\
    CACNF (Ours) & \textbf{56.9} & \textbf{82.5} & \textbf{67.1} & \textbf{90.4} & \textbf{27.1} & \textbf{33.9} & \textbf{37.1} & \textbf{45.5} \\ \bottomrule
\end{tabular}
}
\end{center}
\caption{Something-Else SOTA comparisons. \textbf{Left:} Compositional setting, \textbf{Right:} Few-shot setting. \textbf{Top:} Layout-based methods, \textbf{Bottom:} I3D and Multimodal methods.}
\label{table:compositional-few-shot}
\end{table*}

\textbf{Compositional action recognition}. We report results in Table~\ref{table:compositional-few-shot} (Left). In both the obj. predictions and oracle setting, we observe that STLT outperforms STIN and the other methods, with a more prominent difference in performance in the oracle setting. We also observe that STLT's performance is remarkably close to the best multimodal concurrent method in the oracle setting -- SFI, an indication that, if perfect object detections are available, MHA captures finer interactions between the objects compared to a (convolutional) GNN, 1D convolution, etc. In the multimodal section -- Table~\ref{table:compositional-few-shot} (bottom), obj. predictions setting, CACNF outperforms the other methods significantly, suggesting a noteable improvement in robustness w.r.t. compositional data (a likely real-life scenario).\par

\textbf{Few-shot action recognition.} We report results in Table~\ref{table:compositional-few-shot} (Right). We observe that in the obj. predictions setting, STLT outperforms STIN in both the 5-shot and 10-shot setup. In the oracle setting, STLT significantly outperforms the other methods as per the top-1 accuracy, allowing a significant gap for improvement as object detectors continue to improve \cite{dai2021dynamic}. Furthermore, CACNF outperforms the other multimodal methods in the obj. predictions setting, with a bigger difference in top-1 acc. in the 10-shot setup. We interpret the performance improvement as evidence that STLT and CACNF can successfully generalize in a low-data regime, a valuable observation considering the cost of acquiring curated video data.\par

\subsection{Action Genome: Coping with background-cluttered videos}
Object-centric layout-based models appear to be unsuited for dealing with background-cluttered videos. To address such concerns, we use Action Genome \cite{ji2020action}, as it consists of videos
\begin{inparaenum}[(i)]
\item of people performing actions at home, where it is hard to isolate the objects specific to the action,
\item with objects presence overlapping between training and testing.
\end{inparaenum}
We measure action recognition mAP, as well as mAP relative improvement on top of trained I3D \cite{carreira2017quo}, by ensembling it with STLT. We conduct experiments in a setup where we replace all specific categories, e.g., book, broom, phone, etc., except for person, with a generic ``object'' category, therefore having only 2 object categories (person, object), and the default setup, where we utilize all 38 categories as input to STLT. We compare against:
\begin{inparaenum}[(i)]
\item \textbf{LFB} \cite{wu2019long}: Long-term feature bank model, which performs well when video features are aggregated over time;
\item \textbf{SGFB} \cite{ji2020action}: Method which predicts (or uses ground truth in oracle setting) symbolic scene graph, further encoded and combined with LFB \cite{wu2019long}.
\end{inparaenum}

\begin{wraptable}{R}{7cm}
\vspace{-20pt}
\begin{center}
\resizebox{0.5\textwidth}{!}{
\begin{tabular}{ccccc} \toprule
    {Method} & {Input modalities} & Num. categories & {Obj. predictions mAP} & {Oracle mAP} \\ \midrule
    LFB \cite{wu2019long} & Video Appearance & \NA & 42.5 & 42.5 \\
     SGFB \cite{ji2020action} & Scene Graphs \& Video Appearance & 38 & 44.3 (1.8) & 60.3 (17.8) \\ \midrule
     I3D (Ours) \cite{carreira2017quo}
     & Video Appearance & \NA & 33.5 & 33.5 \\ \midrule
    STLT (Ours) & Obj. detections & 2 & 16.1 & 19.9 \\
    STLT + I3D (Ours) & Obj. detections \& Video Appearance & 2 & 33.8 (0.3) & 36.5 (3.0) \\ \midrule
    STLT (Ours) & Obj. detections & 38 & 30.2 & 60.6 \\
    STLT + I3D (Ours) & Obj. detections \& Video Appearance & 38 & 38.5 (5.0) & 61.63 (28.1) \\ 
    \bottomrule
\end{tabular}
}
\end{center}
\caption{Action Genome results. \textbf{From top to bottom:} Baselines, I3D, STLT and STLT + I3D ensemble with 2 generic obj.~categories (person, obj.), STLT and STLT + I3D ensemble with all 38 obj.~categories (person, book, phone, etc.). Relative mAP improvement over appearance method in parenthesis.}
\label{table:charades}
\end{wraptable}
In Table~\ref{table:charades}, when only 2 object categories (person, object) are registered in STLT, we observe that STLT yields significantly weaker performance compared to a standard appearance method, e.g., I3D. Interestingly, despite the negatively biased setup, i.e., the actions in Action Genome are object-specific, e.g., opening a book, while the input categories are object-agnostic -- person/object, STLT still boosts I3D's performance by 3.0 mAP points in the oracle setting. When all 38 object categories are registered in STLT, we observe that STLT performs well even in the obj. predictions setting, considering the Faster R-CNN's $\sim$ 11.5 average precision (AP) on the validation set. In the oracle setting the performance increases drastically, being on par with SGFB (which relies on ground truth scene graph), indicating a high upper bound, considering that object detection is merely a subset of scene graph generation. In the obj. predictions setting, we observe a solid relative improvement over I3D by ensembling STLT with I3D, even larger compared to SGFB and LFB\footnote{The appearance model performance and the relative improvement are most likely inversely proportional.}, concluding that STLT reasonably copes with background clutter, and successfully boosts the performance of an appearance model -- I3D.

\section{Conclusion}
In this paper we shed light on the problem of compositional and few-show action recognition. We advocated the use of multi-head attention over spatio-temporal layouts, and attempted to reach a conclusion how layout- and appearance-based models should be fused. Our main empirical findings suggest that
\begin{inparaenum}[(i)]
\item a layout-based model is robust w.r.t. compositional data, and generalizes from a few samples,
\item when fusing a layout- and appearance-based model, it is crucial for the individual models to preserve their capabilities,
\item even on non-compositional, background cluttered video datasets, a layout-based model can reasonably recognize human actions, and boosts the performance of appearance-based models.
\end{inparaenum}\par
A limitation which remains is that layout-based models are highly dependent on the object detections quality. Notice, however, the high upper bound (oracle setting), combined with the fact that for compositional action recognition, the requirement is class-agnostic object detections. Lastly, by relying on an object detector the overall model complexity increases, which is detrimental to the speed. Ideally, an off-the-shelf appearance based model should exhibit object-centric reasoning abilities, which we leave for future work.

\section*{Acknowledgements}
We acknowledge funding from the Flemish Government under the Onderzoeksprogramma Artifici\"{e}le Intelligentie (AI) Vlaanderen programme. We also thank Dina Trajkovska for the help with the figures.

\bibliography{main}
\end{document}